\documentclass{opt2021} % Include author names
\title[Gaussian Graphical Models as an Ensemble Method]{Gaussian Graphical Models as an Ensemble Method for Distributed Gaussian Processes}

\optauthor{%
\Name{Hamed Jalali} \Email{hamed.jalali@wsii.uni-tuebingen.de}\\
\Name{Gjergji Kasneci} \Email{gjergji.kasneci@uni-tuebingen.de}\\
\addr University of Tuebingen, Germany}

\begin{document}

\maketitle

\begin{abstract}%
Distributed Gaussian process (DGP) is a popular approach to scale GP to big data which divides the training data into some subsets, performs local inference for each partition, and aggregates the results to acquire global prediction. To combine the local predictions, the \emph{conditional independence assumption} is used which basically means there is a perfect diversity between the subsets. Although it keeps the aggregation tractable, it is often violated in practice and generally yields poor results. In this paper, we propose a novel approach for aggregating the Gaussian experts' predictions by Gaussian graphical model (GGM) where the target aggregation is defined as an unobserved latent variable and the local predictions are the observed variables. We first estimate the joint distribution of latent and observed variables using the Expectation-Maximization (EM) algorithm. The interaction between experts can be encoded by the precision matrix of the joint distribution and the aggregated predictions are obtained based on the property of conditional Gaussian distribution. Using both synthetic and real datasets, our experimental evaluations illustrate that our new method outperforms other state-of-the-art DGP approaches.
\end{abstract}

\section{Introduction}

 Gaussian processes (GPs) are powerful non-parametric statistical methods based on Bayes' theorem. Without the need for restrictive assumptions, they are capable to estimate complex models with a low amount of uncertainty. They have been widely used in practice, e.g. optimization \cite{Shahriari}, data visualization \cite{Lawrence}, reinforcement learning \cite{Deisenroth2013}, multitask learning \cite{Alvarez}, online streaming models \cite{Le}, and time series analysis \cite{ Tobar}. Despite many advantages, GPs suffer from their computational costs where they poorly scale with the size of the dataset. 
The prominent distributed Gaussian processes (also called local approximation GPs) are based on the divide-and-conquer approach. It means the training data is divided into some partitions (called experts), the local inference is done for each partition separately, and at the end, these local estimations are combined using an ensemble method. All experts share the same hyper-parameters, which leads to automatic regularisation and the model tends to prevent the overfitting of individual experts \cite{Deisenroth}.

In a DGP, the \emph{conditional independence} assumption (CI) between partitions allows factorizing the global posterior distribution as a product of local distributions. Although this assumption reduces the computational cost, it is often violated in practice. However, solutions that deal with the dependency problem (e.g. NPAE method \cite{Rulliere}) suffer from extra computational costs and therefore, are impractical for large data sets. 

The key contribution of our work lies in aggregating the local experts' predictions considering their dependencies. Unlike conventional DGPs, here the CI assumption is violated to improve the prediction quality. The conditional dependency is inferred as the interactions between nodes in a continuous form of a Markov random field (MRF). We consider the local and latent experts as nodes of an undirected graph. Then, the Gaussian graphical model (GGM) is used to construct the undirected graph between Gaussian experts and their interactions. Since the latent expert is unobserved, we use the latent variable Gaussian graphical model (LVGGM) to estimate the joint distribution of observed and latent experts. The final predictions are the mean of the conditional distribution of the latent expert given observed experts. Relative to the available baselines, our approach substantially provides competitive prediction performance than other state-of-the-art (SOTA) approaches, which use the CI assumption.
The structure of the paper is as follows. Section \ref{problem_set_up} introduces the problem formulation and related works. In Section \ref{proposed_method} the proposed model and the inference process are presented. Section \ref{experiments} shows the experimental results and we conclude in Section \ref{conclusion}.

\section{Background and Problem Set-up} \label{problem_set_up}

\subsection{Background}
Let us consider the regression problem $y=f(x)+\epsilon$, where $x\in R^d$ and $\epsilon \sim \mathcal{N}(0,\sigma^2)$, and the Gaussian likelihood is $p(y|f)=\mathcal{N}(f, \sigma^2 I)$. The objective is to learn the latent function \textit{f} from a training set $\mathcal{D}=\{X,y\}$ of size $n$. The Gaussian process regression is a collection of random variables of which any finite subset has a joint Gaussian distribution. The GP then describes a prior distribution over the latent functions as $f \sim GP\left(0,k(x,x') \right)$, where $k(x,x')$ is the covariancee function (kernel) with hyperparameters $\psi$. To train the GP, the hyperparameters $\theta=\{\sigma^2, \psi\}$ should be determined such that they maximise the log-marginal likelihood,
\begin{equation} \label{eq:1}
\log p(y|X)=-\frac{1}{2}y^T\mathcal{C}^{-1}y - \frac{1}{2} \log|\mathcal{C}|- \frac{n}{2} \log2\pi,
\end{equation}
where $\mathcal{C}=K+\sigma^2I$ and $K=k(X,X)$. According to \eqref{eq:1}, the training step scales as $\mathcal{O}(n^3)$ because it is affected by the inversion and determinant of the $n \times n$ matrix $\mathcal{C}$. Therefore, for large data sets, GP training is a time-consuming task and imposes limitations on the scalability of GPs.

\subsection{Distributed Gaussian Process}
The term distributed Gaussian process was proposed by \cite{Deisenroth} uses the fact that the computations of the standard GP can be distributed among individual computing units. To do that, one divides the full training data set $\mathcal{D}$ into $M$ partitions (called experts) and trains standard GPs on these partitions. Let  $\mathcal{D}^{'}= \{\mathcal{D}_1,\ldots,\mathcal{D}_M\}$ be the partitions, and $X_i$ and $y_i$ be the input and output of partition $\mathcal{D}_i$. All GP experts are trained jointly and share a single set of hyper-parameters $\theta=\{\sigma^2,\psi\}$. For a test set $X^*$ of size $n_t$, the local prediction of the $i$-th GP expert $\mathcal{M}_i$ is:
\begin{align}
\mu_i^* &= k_{i*}^T(K_i+\sigma^2I)^{-1}y_i,\label{eq:2}
\end{align}
where $K_i=k(X_i,X_i)$, and $k_{i*}=k(X_i,X^*)$. 

Aggregating the experts in DGP is based on the assumption that they are conditionally independent. For a test input $x^*$, the posterior distribution of DGP is given as the product of multiple local densities, i.e. $p(y^*|\mathcal{D},x^*) \propto \prod_{i=1}^M p_i(y^*|\mathcal{D}_i,x^*)$. 
The most popular aggregations are generalised product of experts (GPoE) \cite{Cao}, robust Bayesian committee machine (RBCM) \cite{Deisenroth} and generalized robust Bayesian committee machine (GRBCM) \cite{Liu2018}, see Appendix \ref{App.DGP}.

\subsection{Dependency}
The CI assumption is used widely in ensemble methods for both regression and classification problems \cite{Moreira,Parisi}. The DGPs use CI to reduce the computational costs of the training and prediction processes. However, their predictions are not accurate enough and CI-based aggregation generally returns sub-optimal solution \cite{Jaffe,jalali2020}. In local approximation GPs, the dependency between experts has been discussed in few works. For instance, the nested pointwise aggregation of experts (NPAE) method \cite{Rulliere} uses the internal correlation between local experts and the dependency between local experts and target variable $y^*$. However, this pointwise aggregation suffers from high time complexity which cubically depends on the number of experts at each test point, i.e. $\mathcal{O}(n_t M^3)$, and therefore, it is not an efficient solution for large datasets. Figure \ref{aggregation_graphs} in Appendix \ref{App.graph} shows the computational graphs of CI-based and dependency-based aggregation strategies.

\section{Aggregating Conditionally dependent Experts with an Undirected Graph} \label{proposed_method}

At the heart of our work is the following ingredient. First, we assume that $y_i$ in \eqref{eq:2} has not yet been observed, see \cite{Rulliere}. Then the experts' predictions $\mu^*_i$ can be considered as a \emph{random variable}. This allows us to leverage correlations between experts. Then, we exert the Gaussian graphical model, where the nodes of the graph are the experts (local and latent) and the edges are the interactions between them.

\subsection{Aggregating Dependent Experts' Predictions}
Assume the Gaussian experts $\mathcal{M}=\{\mathcal{M}_1,\ldots,\mathcal{M}_M\}$ have been trained on separated subsets and let $\mu^*=[\mu_1^*,\ldots,\mu_M^*]^T$ be a $ n_t \times M$ matrix that contains their centered predictions at $n_t$ test points. As a consequence of the choice of the prior, the joint distribution of the local experts $\mu^*$ and target expert $y^*$ is multivariate Gaussian distribution because any vector of linear combinations of observation is itself a Gaussian vector. Let $\Sigma_{y^*\mu^*}$ encodes the correlation between latent expert $y^*$ and local experts $\mu^*$, and $\Sigma_{\mu^* \mu^*}$ depicts the correlation between local experts. Employing the properties of conditional Gaussian distributions for the centered random vector allows for the following aggregation:
\begin{equation} \label{aggregated_estimator}
\begin{split}
 y_A^* =\Sigma_{y^*\mu^*}^T \Sigma_{\mu^* \mu^*}^{-1}\mu^*.   
\end{split}
\end{equation}
which is the mean of conditional distribution of $y^*$ given $\mu^*$, i.e. $p(y^*|\mu^*)$.

\begin{proposition}[\textbf{BLUP}]
$y_A^*$ is the best linear unbiased predictor of $y^*$, i.e. for linear estimators of the form $\beta \mu^* = \sum_{i=1}^M \beta_i \mu_i^*$, the mean square error $(y^*- \beta \mu^*)^2$ is minimized when $\beta=\Sigma_{y^*\mu^*}^T \Sigma_{\mu^*\mu^*}^{-1}$.
\label{prop_blup}
\end{proposition}

The proof of Proposition \ref{prop_blup} can be found in Appendix \ref{App.proof}. In the next subsection, we show how the GGM can be adapted to the local approximation problem with a latent target variable and suggest a new method to compute the aggregated estimator $y^*_A$.

\subsection{Gaussian Graphical Models for Dependent Gaussian Experts}
Gaussian graphical models \cite{Rue,Uhler,Drton} are continuous forms of pairwise MRFs which assume the variables in the network follow a multivariate Gaussian distribution. The distribution for a GGM is 

\begin{equation} \label{GGM_variance}
\begin{aligned}
 p(\mu^*|\xi, \Omega) \propto \exp\left\{-\frac{1}{2}(\mu^*-\xi)^T\Omega (\mu^*-\xi) \right\},
\end{aligned}
\end{equation}

where $\mu^*=\{\mu^*_1,\ldots,\mu^*_M\}$ are the experts,  and $\xi$ and $\Omega$ are the mean and precision, respectively. The matrix $\Omega$ is also known as the potential or information matrix. In a GGM, if $\Omega_{ij}=0$, then $\mu^*_i$ ard $\mu^*_j$ are conditionally independent given all other variables, i.e. there is no edge between $\mu^*_i$ ard $\mu^*_j$ in the graph. GGMs use the common sparsity assumption, that is, there are only few edges
in the network and thus the precision matrix is sparse. To this end, the graphical Lasso (GLasso) regression \cite{Friedman2008} is used to perform neighborhood selection for the network. It maximizes the log-likelihood subject to an element-wise $\mathcal{L}_1$ norm penalty on $\Omega$. Precisely, for sample covariance S and Gaussian log-likelihood $ \mathcal{L}(\Omega; S)=\log |\Omega| - trace(S \; \Omega)$, the objective function is 
\begin{equation}  \label{GLasso}
 \widehat{\Omega}_{\lambda}= \arg\min_{\Omega} \; \left( -\mathcal{L}(\Omega; S) + \lambda \left\Vert \Omega \right\Vert_1 \right).
\end{equation}
The precision matrix $\Omega$ has been used before to find clusters of strongly dependent experts \cite{jalali2020} and selecting most important experts in local approximation \cite{jalali2021}.

\subsection{GGM-Based Aggregation using EM Algorithm}
The main input of the GLasso method is the sample covariance of our observations. Since the targeted expert $y^*$ is unobserved, one row (column) in S, related to $y^*$ is unknown. Let $S_{\mu^* \mu^*}$ is a known $M \times M$ matrix of the sample covariance of the observed variables $\mu^*$, $S_{y^*\mu^*}$ is an unknown $1 \times M$ vector that shows the sample covariance between latent and observed expert, and $S_{y^* y^*}$ is the internal potential of a latent expert. To use the GLasso, it is needed to estimate unknown partitions of $S$, i.e. $S_{y^*\mu^*}$ and $S_{y^* y^*}$. Here, we explain how the expected-maximization algorithm can help us.
\paragraph{\textbf{E-Step:}} The E-step Calculates $Q(\Omega \mid \Omega^{(t)})$, the expected value of the penalized negative log-likelihood function with respect to the conditional distribution of $y^*$ given $\mu^*$ under the current estimate $\Omega^{(t)}$ of $\Omega$:
\begin{align*}
Q(\Omega \mid \Omega^{(t)})=\operatorname {E} _{y^* \mid  {\mu^*} ,\Omega^{(t)}} &{}\left[ -\mathcal{L}(\Omega; S) + \lambda \left\Vert \Omega \right\Vert_1 \right] 
= -\log |\Omega| + trace\{ E_{y^*|\mu^*,\Omega^{(t)}}(S)\Omega\} + \lambda \left\Vert \Omega \right\Vert_1 .
\end{align*}
Let $\Sigma^{(t)}=(\Omega^{(t)})^{-1}$, the conditional distribution of $y^*$ given $\mu^*$ under the current estimate $\Omega^{(t)}$ follows   
\[ N\left(\Sigma^{(t)}_{y^* \mu^*} (\Sigma^{(t)}_{\mu^* \mu^*})^{-1} \mu^*, \Sigma^{(t)}_{y^*} - \Sigma^{(t)}_{y^* \mu^*} (\Sigma^{(t)}_{\mu^* \mu^*})^{-1} \Sigma^{(t)}_{\mu^* y^*} \right).\]
Therefore, unknown partitions of $\widehat{S}=E_{y^*|\mu^*,\Omega^{(t)}}(S)$ can be estimated as below: 
\begin{equation} \label{eq:S_y_mu}
\widehat{S}_{\mu^*y^*}=S_{\mu^* \mu^*} (\Sigma^{(t)}_{\mu^* \mu^*})^{-1}\Sigma^{(t)}_{\mu^* y^*},
\end{equation}
\begin{multline} \label{eq:S_y_y}
\widehat{S}_{y^* y^*}=\Sigma^{(t)}_{y^* y^*} - \Sigma^{(t)}_{y^* \mu^*} (\Sigma^{(t)}_{\mu^* \mu^*})^{-1} \Sigma^{(t)}_{\mu^* y^*} + \Sigma^{(t)}_{y^* \mu^*} (\Sigma^{(t)}_{\mu^* \mu^*})^{-1} S_{\mu^* \mu^*} (\Sigma^{(t)}_{\mu^* \mu^*})^{-1} \Sigma^{(t)}_{\mu^* y^*}.
\end{multline}
\vskip -0.5in
\paragraph{\textbf{M-Step}}:  This step returns the updated precision matrix $\displaystyle {\boldsymbol {\Omega }}^{(t+1)}$ that maximize $Q(\Omega \mid \Omega ^{(t)})$ over all $(M+1)\times(M+1)$ positive-definite matrices $\Omega$. It is a \textit{GLasso} problem and is equivalent to this minimization problem:
\begin{equation}  \label{EM_maximization}
 \Omega^{(t)}=\underset{\Omega }{\arg\min} \left(- \log |\Omega| + trace\{\widehat{S}\Omega\} + \lambda \left\Vert \Omega \right\Vert_1 \right).
\end{equation}
Algorithm \ref{alg:aggregation} summarizes the whole procedure of the proposed ensemble, EMGGM. 

\begin{algorithm2e}
 \caption{GGM-Based Experts Aggregation (EMGGM)}
 \label{alg:aggregation}
 \SetAlgoLined
  \KwData{$\mu^*$, $\lambda$, $R$ (number of iterations)}
  \KwResult{Aggregated predictions $y_A^*$  }
  Initialize $y^*$\;  
  Calculate sample covariance $S^{(0)}$ of $(y^*,\mu^*)$\;
  Estimate the initial parameter $\Omega^{(0)}$ using Equation \eqref{GLasso}\;
  $t\gets 1$\;
  \While{$t \leq R$}{
  Estimate $E_{y^*|\mu^*,\Omega^{(t)}}(S_{\mu^* y^*})$ using Equation \eqref{eq:S_y_mu} \;
  Estimate $E_{y^*|\mu^*,\Omega^{(t)}}(S_{y^* y^*})$ using Equation \eqref{eq:S_y_y}\;
  Update the sample covariance as  $S^{(t)}=E_{y^*|\mu^*,\Omega^{(t)}}(S)$\;
  Update the precision matrix $\Omega^{(t)}$ using Equation \eqref{EM_maximization} \;
  $\Sigma^{(t)} \gets (\Omega^{(t)})^{-1}$ and $t\gets t+1$\;
  } 
  Estimate the aggregated prediction $y_A^*$ using Equation \eqref{aggregated_estimator} \;
\end{algorithm2e}

\subsection{Discussion and Challenges}

The proposed ensemble is capable to aggregate local experts considering their dependencies and its computational and storage costs are much smaller than NPAE which uses dependent experts, see Appendix \ref{App.complexity}. Besides, the normality assumption for joint distribution is not a restrictive assumption and can be relaxed, see Appendix \ref{App.Gaussian}. This gives the result that the proposed strategy can aggregate non-Gaussian experts. 

An EM iteration does increase the likelihood function $\mathcal{L}(\Omega; S)$. However, no guarantee exists that the sequence converges to a maximum likelihood estimator. It is only guaranteed to converge to a point with zero gradient with respect to the parameters. So it can indeed get stuck at saddle points, see \cite{McLachlan}. The converges property of the EM algorithm can be improved using a variety of heuristic or meta-heuristic approaches that enable EM to escape a local maximum, e.g. hill climbing and simulated annealing.

To avoid challenges in the convergence of EM, latent variable GGM (LVGGM) can be used. Maximizing the likelihood function of an LVGGM leads to a nonlinear optimization problem which is solved by convex or non-convex optimization methods. This strategy can be studied in future works to estimate the aggregated estimator $y_A^*$ in \eqref{aggregated_estimator}, see Appendix \ref{App.LVGGM}.
 
\section{Experiments} \label{experiments}
In this section, we evaluate the prediction quality of the aggregated estimator using the conventional mean absolute error (MAE) and the root mean squared error (RMSE). 
We use the simulated data of a one-dimensional analytical function \cite{Liu2018,jalali2020},
\begin{equation}
f(x) = 5x^2sin(12x) + (x^3 -0.5)sin(3x-0.5)+4cos(2x) + \epsilon,  \label{f_x}
\end{equation}

where $\epsilon \sim \mathcal{N}\left(0, (0.2)^2\right)$. We generate $n=10^4$ training points in $[0,1]$, and $n_t=10^3$ test points in $[-0.2,1.2]$. The data is normalized to zero mean and unit variance. We vary the number of experts, $M=\{10,20,30,40\}$, to evaluate different partition sizes. The prediction quality of the proposed ensemble is compared with the other baselines: GPoE \cite{Cao}, RBCM \cite{Deisenroth}, GRBCM \cite{Liu2018}, NPAE \cite{Rulliere}, and the full GP. 
We use the standard squared exponential kernel, a Gaussian likelihood and the $K$-means partitioning method. 

\begin{figure}
\centering
\subfigure[MAE]{\includegraphics[width=0.32\textwidth]{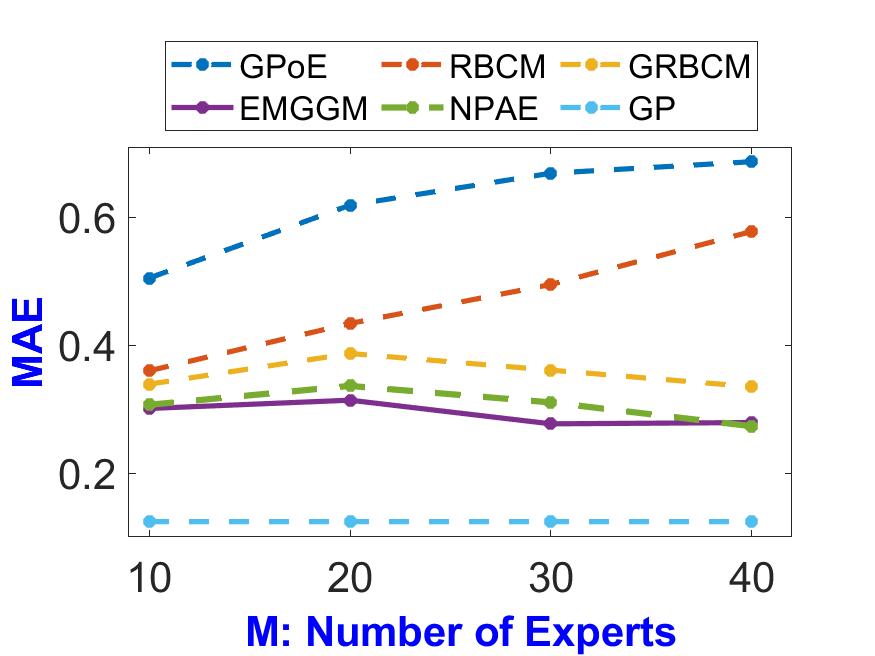}}
\subfigure[RMSE]{\includegraphics[width=0.32\textwidth]{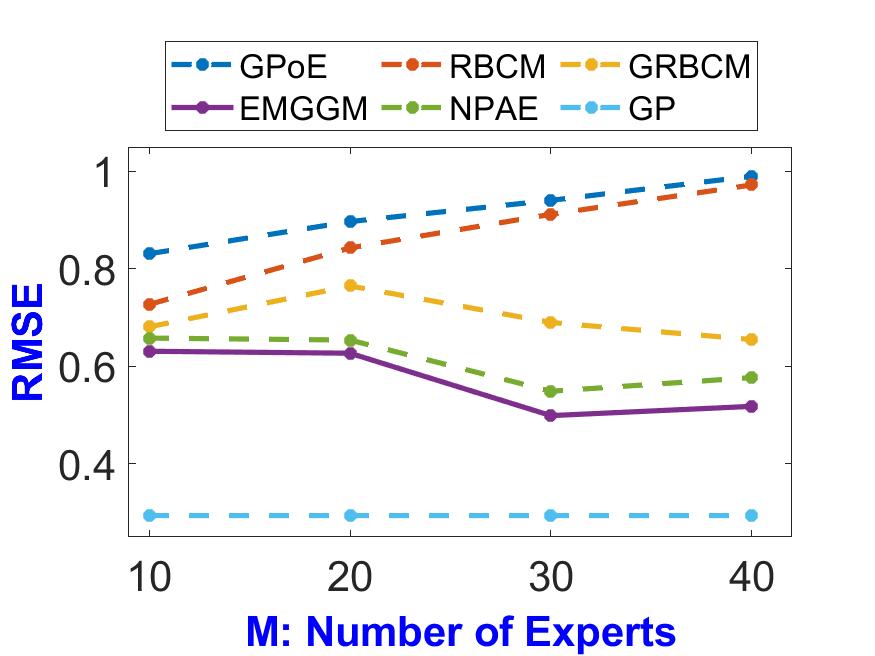}} 
\subfigure[Log(Time)]{\includegraphics[width=0.32\textwidth]{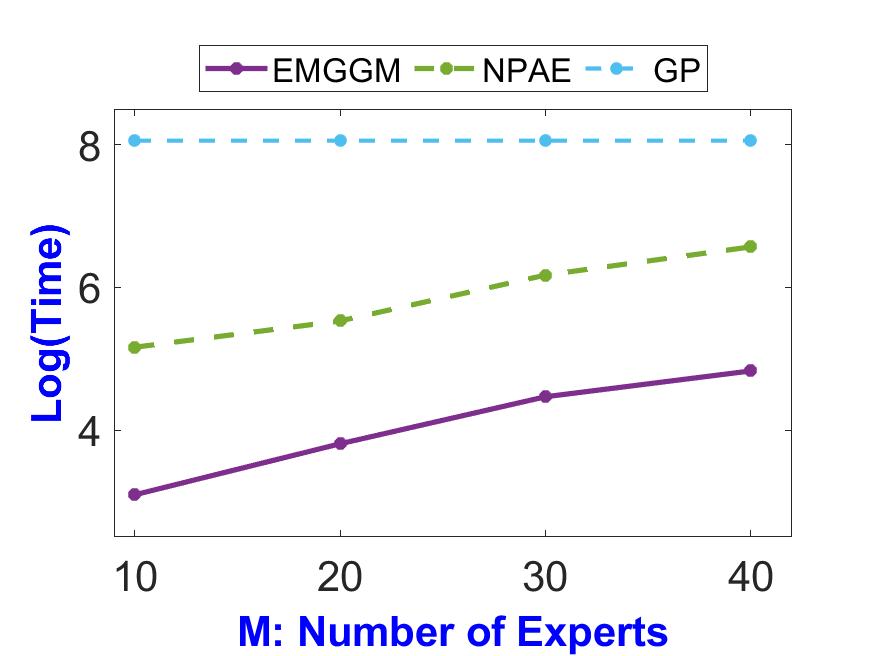}} 
\caption{\textbf{Prediction quality} of DGP methods with respect for different number of experts M.}
\label{Prediction_Quality}
\vskip -0.2in
\end{figure}

Figure \ref{Prediction_Quality} (a) and Figure \ref{Prediction_Quality} (b) depict the prediction quality of different baselines. The ensemble methods that use dependency between experts, i.e. EMGGM and NPAE, outperform the CI-based baselines. However, the proposed method has slightly better predictions than NPAE. Figure \ref{Prediction_Quality} (c) presents the computation time of the ensembles that use dependency between experts. Remarkably, EMGGM provides predictions in just a fraction of NPAE's running time.

\section{Conclusion}\label{conclusion}
In this work, we have proposed a novel ensemble method, EMGGM, for distributed GPs which  aggregate dependent local experts' predictions using GGMs. Our proposed approach uses undirected graphical models and EM algorithm to estimate the final predictions. Through empirical analyses, we illustrated the superiority of EMGGM over existing SOTA aggregation methods. Finally, we hope to use our insights to develop aggregations that provide the full predictive distribution.

\bibliography{sample}

\begin{thebibliography}{32}
\providecommand{\natexlab}[1]{#1}
\providecommand{\url}[1]{\texttt{#1}}
\expandafter\ifx\csname urlstyle\endcsname\relax
  \providecommand{\doi}[1]{doi: #1}\else
  \providecommand{\doi}{doi: \begingroup \urlstyle{rm}\Url}\fi

\bibitem[Alvarez et~al.(2012)Alvarez, Rosasco, and Lawrence]{Alvarez}
M.~A. Alvarez, L.~Rosasco, and N.~D. Lawrence.
\newblock Kernels for vector-valued functions: A review.
\newblock \emph{Foundations and Trends in Machine Learning}, 4\penalty0
  (3):\penalty0 195--266, 2012.

\bibitem[Candès et~al.(2011)Candès, Li, y.~MA, and j.~Wright]{Candes}
E.J. Candès, X.~Li, y.~MA, and j.~Wright.
\newblock Robust principal component analysis?
\newblock \emph{Journal of the ACM (JACM)}, 58\penalty0 (3):\penalty0 1--37,
  2011.

\bibitem[Cao and Fleet(2014)]{Cao}
Y.~Cao and D.~J. Fleet.
\newblock Generalized product of experts for automatic and principled fusion of
  gaussian process predictions.
\newblock \emph{arXiv preprint arXiv:1410.7827}, 2014.

\bibitem[Chandrasekaran et~al.(2012)Chandrasekaran, Parrilo, and
  Willsky]{Chandrasekaran}
V.~Chandrasekaran, P.A. Parrilo, and A.~S. Willsky.
\newblock Latent variable graphical model selection via convex optimization.
\newblock \emph{The Annals of Statistics}, 40:\penalty0 1935--1967, 2012.

\bibitem[Deisenroth and Ng(2015)]{Deisenroth}
M.~P. Deisenroth and J.~W. Ng.
\newblock Distributed gaussian processes.
\newblock \emph{International Conference on Machine Learning}, pages
  1481--1490, 2015.

\bibitem[Deisenroth et~al.(2013)Deisenroth, Fox, and Rasmussen]{Deisenroth2013}
M.~P. Deisenroth, D.~Fox, and C.~E. Rasmussen.
\newblock Gaussian processes for data-efficient learning in robotics and
  control.
\newblock \emph{IEEE Transactions on Pattern Analysis and Machine
  Intelligence}, 37\penalty0 (2):\penalty0 408--423, 2013.

\bibitem[Drton and Maathuis(2017)]{Drton}
M.~Drton and M.H. Maathuis.
\newblock Structure learning in graphical modeling.
\newblock \emph{Annual Review of Statistics and Its Application}, 4:\penalty0
  365--393, 2017.

\bibitem[Friedman et~al.(2008)Friedman, Hastie, and Tibshirani]{Friedman2008}
J.~Friedman, T.~Hastie, and R.~Tibshirani.
\newblock Sparse inverse covariance estimation with the graphical lasso.
\newblock \emph{Biostatistics}, 9\penalty0 (3):\penalty0 432--441, 2008.

\bibitem[Jaffe et~al.(2016)Jaffe, Fetaya, Nadler, Jiang, and Kluger]{Jaffe}
A.~Jaffe, E.~Fetaya, B.~Nadler, T.~Jiang, and Y.~Kluger.
\newblock Unsupervised ensemble learning with dependent classifiers.
\newblock \emph{In Artificial Intelligence and Statistics}, pages 351--360,
  2016.

\bibitem[Jalali and Kasneci(2020)]{jalali2020}
H.~Jalali and G.~Kasneci.
\newblock Aggregating dependent gaussian experts in local approximation.
\newblock \emph{arXiv preprint arXiv:2010.08873}, 2020.

\bibitem[Jalali et~al.(2021)Jalali, Pawelczyk, and Kasneci]{jalali2021}
H.~Jalali, M.~Pawelczyk, and G.~Kasneci.
\newblock Gaussian experts selection using graphical models.
\newblock \emph{arXiv preprint arXiv:2102.01496}, 2021.

\bibitem[Lafferty et~al.(2012)Lafferty, Liu, and Wasserman]{Lafferty}
J.~Lafferty, H.~Liu, and L.~Wasserman.
\newblock Sparse nonparametric graphical models.
\newblock \emph{Statistical Science}, 27\penalty0 (4):\penalty0 519–537,
  2012.

\bibitem[Lawrence(2005)]{Lawrence}
N.~Lawrence.
\newblock Taking the human out of the loop: A review of bayesian optimization.
\newblock \emph{Journal of Machine Learning Research}, 6:\penalty0 1783--1816,
  2005.

\bibitem[Le et~al.(2017)Le, Nguyen, Nguyen, Nguyen, and Phung]{Le}
T.~Le, K.~Nguyen, V.~Nguyen, T.~D. Nguyen, and D.~Phung.
\newblock Gogp: Fast online regression with gaussian processes.
\newblock \emph{IEEE International Conference on Data Mining}, pages 257--266,
  2017.

\bibitem[Li and Solea(2018)]{li}
B.~Li and E.~Solea.
\newblock A nonparametric graphical model for functional data with application
  to brain networks based on fmri.
\newblock \emph{Journal of the American Statistical Association}, 113\penalty0
  (524):\penalty0 1637–1655, 2018.

\bibitem[Liu et~al.(2009)Liu, Lafferty, and Wasserman]{Liu2009}
H.~Liu, J.~Lafferty, and L.~Wasserman.
\newblock nonparanormal: Semiparametric estimation of high dimensional
  undirected graphs.
\newblock \emph{The Journal of Machine Learning Research (JMLR)}, 10:\penalty0
  2295–2328, 2009.

\bibitem[Liu et~al.(2018)Liu, Cai, Ong, and Wang]{Liu2018}
H.~Liu, J.~Cai, Y.~Ong, and Y.~Wang.
\newblock Generalized robust bayesian committee machine for large-scale
  gaussian process regression.
\newblock \emph{International Conference on Machine Learning}, pages 1--10,
  2018.

\bibitem[McLachlan and Krishnan(2008)]{McLachlan}
G.~J. McLachlan and T.~Krishnan.
\newblock \emph{The EM Algorithm and Extensions}.
\newblock Wiley-Interscience, 2 edition, 2008.

\bibitem[Mendes-Moreira et~al.(2012)Mendes-Moreira, Soares, Jorge, and
  Sousa]{Moreira}
J.~Mendes-Moreira, C.~Soares, A.M Jorge, and J.F.D Sousa.
\newblock Ensemble approaches for regression: A survey.
\newblock \emph{Acm computing surveys (csur)}, 45\penalty0 (4):\penalty0 1--40,
  2012.

\bibitem[Mulgrave and Ghosal(2020)]{Mulgrave_2020}
J.~J. Mulgrave and S.~Ghosal.
\newblock Bayesian inference in nonparanormal graphical models.
\newblock \emph{Bayesian Analysis}, 15\penalty0 (2):\penalty0 449--475, 2020.

\bibitem[Parisi et~al.(2014)Parisi, Strino, Nadler, and Kluger]{Parisi}
F.~Parisi, F.~Strino, B.~Nadler, and Y.~Kluger.
\newblock Ranking and combining multiple predictors without labeled data.
\newblock \emph{Proceedings of the National Academy of Sciences}, 111\penalty0
  (4):\penalty0 1253--1258, 2014.

\bibitem[Rue and Held(2005)]{Rue}
H.~Rue and L.~Held.
\newblock \emph{Gaussian Markov random fields: theory and applications}.
\newblock CRC Press, 2005.

\bibitem[Rullière et~al.(2018)Rullière, Durrande, Bachoc, and
  Chevalier]{Rulliere}
D.~Rullière, N.~Durrande, F.~Bachoc, and C.~Chevalier.
\newblock Nested kriging predictions for datasets with a large number of
  observations.
\newblock \emph{Statistics and Computing}, 28\penalty0 (4):\penalty0 849--867,
  2018.

\bibitem[Shahriari et~al.(2016)Shahriari, Swersky, Wang, Adams, and
  Freitas]{Shahriari}
B.~Shahriari, K.~Swersky, Z.~Wang, R.~P. Adams, and N.~De Freitas.
\newblock Taking the human out of the loop: A review of bayesian optimization.
\newblock \emph{Proceedings of the IEEE}, 104\penalty0 (1):\penalty0 148--175,
  2016.

\bibitem[Solea and Dette(2021)]{Solea_2021}
E.~Solea and H.~Dette.
\newblock Nonparametric and high-dimensional functional graphical models.
\newblock \emph{arXiv preprint arXiv:2103.10568s}, 2021.

\bibitem[Tobar et~al.(2015)Tobar, Bui, and Turner]{Tobar}
F.~Tobar, T.~D. Bui, and R.~E. Turner.
\newblock Learning stationary time series using gaussian processes with
  nonparametric kernels.
\newblock \emph{In Advances in Neural Information Processing Systems}, pages
  3501--3509, 2015.

\bibitem[Tresp(2000)]{Tresp}
V.~Tresp.
\newblock A bayesian committee machine.
\newblock \emph{Neural Computation}, 12\penalty0 (11):\penalty0 2719--2741,
  2000.

\bibitem[Uhler(2017)]{Uhler}
C.~Uhler.
\newblock Gaussian graphical models: an algebraic and geometric perspective.
\newblock \emph{arXiv preprint arXiv:1707.04345}, 2017.

\bibitem[Wang et~al.(2016)Wang, Ding, Fang, MacDonald, Sweet, Wang, and
  Chen]{wang_2016}
T.~Wang, Z.~Renand~Y. Ding, Z.~Fang, Z.~Sun M.~L. MacDonald, R.~A. Sweet,
  J.~Wang, and W.~Chen.
\newblock Fastggm: An efficient algorithm for the inference of gaussian
  graphical model in biological networks.
\newblock \emph{PLOS Computational Biology}, 12\penalty0 (2):\penalty0 1--16,
  2016.

\bibitem[Xu et~al.(2017)Xu, Ma, and Gu]{Xu2017}
P.~Xu, J.~Ma, and Q.~Gu.
\newblock Speeding up latent variable gaussian graphical model estimation via
  nonconvex optimizations.
\newblock \emph{Advances in Neural Information Processing Systems}, 2017.

\bibitem[Yuan(2012)]{yuan}
Ming Yuan.
\newblock Discussion: Latent variable graphical model selection via convex
  optimization.
\newblock \emph{The Annals of Statistics}, 40\penalty0 (4):\penalty0
  1968--1972, 2012.

\bibitem[Zhang et~al.(2020)Zhang, Wang, Li, Wang, Chang, and Wang]{zhang_2020}
J.~Zhang, M.~Wang, Q.~Li, S.~Wang, X.~Chang, and B.~Wang.
\newblock Quadratic sparse gaussian graphical model estimation method for
  massive variables.
\newblock \emph{Proceedings of the Twenty-Ninth International Joint Conference
  on Artificial Intelligence}, pages 2964--2972, 2020.

\end{thebibliography}
\newpage
%\clearpage
%\appendix

%\section{More}

\begin{appendices}
\addcontentsline{toc}{section}{Appendices}
%\section*{Appendices.A}

\section{Distributed GP Models}\label{App.DGP}
In this section we review other divide-and-conquer distributed GP approaches in more detail, focusing on how other methods perform expert weighting. There are two main families of distributed GPs: product of experts and Bayesian committee machine. 
\paragraph{Product of Experts.}
The posterior distribution of the PoE model is given by the product of multiple densities (i.e., the experts). Because of the product operation, the prediction quality of PoE suffers considerably from weak experts. To improve on this aspect, \cite{Cao} proposed the GPoE model, which assigns importance weight to the experts.

For independent experts $\{\mathcal{M}\}_{i=1}^M$ trained on different partitions $\mathcal{D}_i$ the predictive distribution for a test input $X^*$ is given by:
\begin{equation} \label{eq:13}
p(y^*|\mathcal{D},X^*)= \prod_{i=1}^M p_i^{\beta_i}(y^*|\mathcal{D}_i,X^*),
\end{equation}
where $\beta = \{\beta_1,\ldots,\beta_M\}$ controls the expert importance. The product distribution in \eqref{eq:13} is proportional to a Gaussian distribution with mean and precision, respectively:

\begin{equation*} 
\mu_{D}^* = \Sigma_{D}^* \sum_{i=1}^M \beta_i(\Sigma_i^*)^{-1}\mu_i^*, \;\;\; 
(\Sigma_{D}^*)^{-1} = \sum_{i=1}^M \beta_i (\Sigma_i^*)^{-1}.  
\end{equation*}
The standard PoE can be recovered by setting $\beta_i=1 ~ \forall i$. The precision corresponding to the PoE prediction, i.e. $(\Sigma_{D}^*)^{-1}$, is a linear sum of individual precision values: hence, an increasing number of local GPs increases the precision and therefore it leads to a decrease in variance, which consequently returns overconfident predictions in areas with little data.

To choose the weights $\beta_i$ in the PoE model several heuristics have been put forward. The authors of \cite{Cao} suggested the difference in differential entropy between the prior and posterior distribution of each expert, i.e. $\beta_i=\frac{1}{2}(\log \Sigma^{**} - \log \Sigma_i^{*})$ where the $(\Sigma^{**})^{-1}$ is the prior precision of $p(y^*)$. This leads to more conservative predictions. To fix this issue, \cite{Deisenroth} suggested to choose simple uniform weights $\beta_i=\frac{1}{M}$, which provides better predictions.
\paragraph{Bayesian Committee Machine.}
The Bayesian committee machine \cite{Tresp} uses the Gaussian process prior $p(y^*)$ for the aggregation step and assumes conditional independence between experts, i.e. $\mathcal{D}_i \perp \!\!\! \perp  \mathcal{D}_j |y^*$ for two experts $i$ and $j$. 
To mitigate the effect of weak experts on aggregation, especially in regions with few data points, \cite{Deisenroth}
proposed the robust Bayesian committee machine (RBCM), which added importance weights $\beta_i$ to the model.
The distributed predictive distribution of this family of models can be written as: 
\begin{equation*} %\label{eq:16}
p(y^*|\mathcal{D},X^*)= \frac{\prod_{i=1}^M p_i^{\beta_i}(y^*|\mathcal{D}_i,X^*)}{p^{\sum_{i=1}^M \beta_i-1}(y^*)}.
\end{equation*}
Its distribution is proportional to a Gaussian distribution with mean and precision, respectively:
\begin{align*} 
\mu_{D}^* = \Sigma_{D}^* \sum_{i=1}^M \beta_i(\Sigma_i^*)^{-1}\mu_i^*, \;\;\;
(\Sigma_{D}^*)^{-1} = \sum_{i=1}^M \beta_i (\Sigma_i^*)^{-1} +(1-\sum_{i=1}^M \beta_i)(\Sigma^{**})^{-1},  %\label{eq:18}
\end{align*}
where the $(\Sigma^{**})^{-1}$ is the prior precision of $p(y^*)$. The general choice of the weights is the difference in differential entropy between the prior $p(y^*|X^*)$ and the posterior $p(y^*|\mathcal{D},X^*)$, i.e. $\beta_i=\frac{1}{2}(\log \Sigma^{**} - \log \Sigma_i^{*})$.

The most recent model in this family is the generalized robust Bayesian committee machine (GRBCM) \cite{Liu2018}. It introduces a base (global) expert and considers the covariance between the base and other local experts. For a global expert $M_b$ and a base partition ${D}_{b}$, the predictive distribution of GRBCM is
\begin{equation}
p(y^*|\mathcal{D},X^*)= \frac{\prod_{i=2}^M p_{bi}^{\beta_i}(y^*|\mathcal{D}_{bi},X^*)}{p_b^{\sum_{i=2}^M \beta_i-1}(y^*|\mathcal{D}_b,X^*)} ,
\end{equation}
where the $p_b(y^*|\mathcal{D}_b,X^*)$ is the predictive distribution of $M_b$, and $p_{bi}(y^*|\mathcal{D}_{bi},X^*)$ is the predictive distribution of an expert trained on the data set $\mathcal{D}_{bi}=\{\mathcal{D}_{b},\mathcal{D}_{i}\}$. The base partition is randomly selected, while the remaining experts can be chosen through a random or disjoint partitioning strategy. It is noteworthy that, for $M$ experts and $m_0$ data points per expert, the GRBCM operates based on $M-1$ experts with $2m_0$ data points per expert. Therein lies the main difference between GRBCM and the other distributed GPs, which use $m_0$ data points per expert only. Since GRBCM assigns more data points to the experts, it trains experts on more informative subsets. 

\section{Computational Graphs of Aggregation Strategies} \label{App.graph}
\begin{figure}[hbt!]
    \centering
    \subfigure[Independent Experts]{\includegraphics[width=0.40\textwidth]{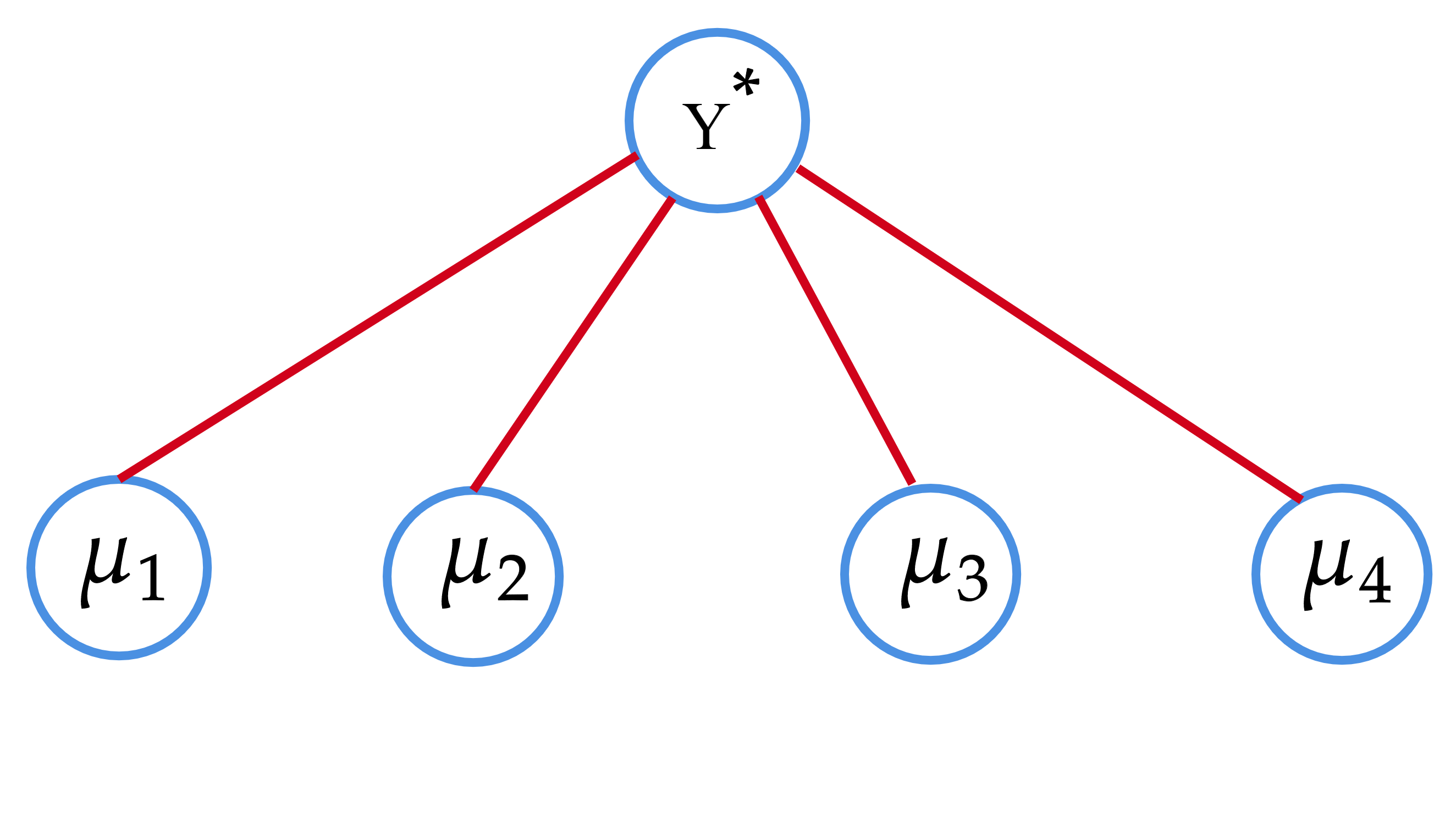}} \hspace{2mm}
    \subfigure[Dependent Experts]{\includegraphics[width=0.40\textwidth]{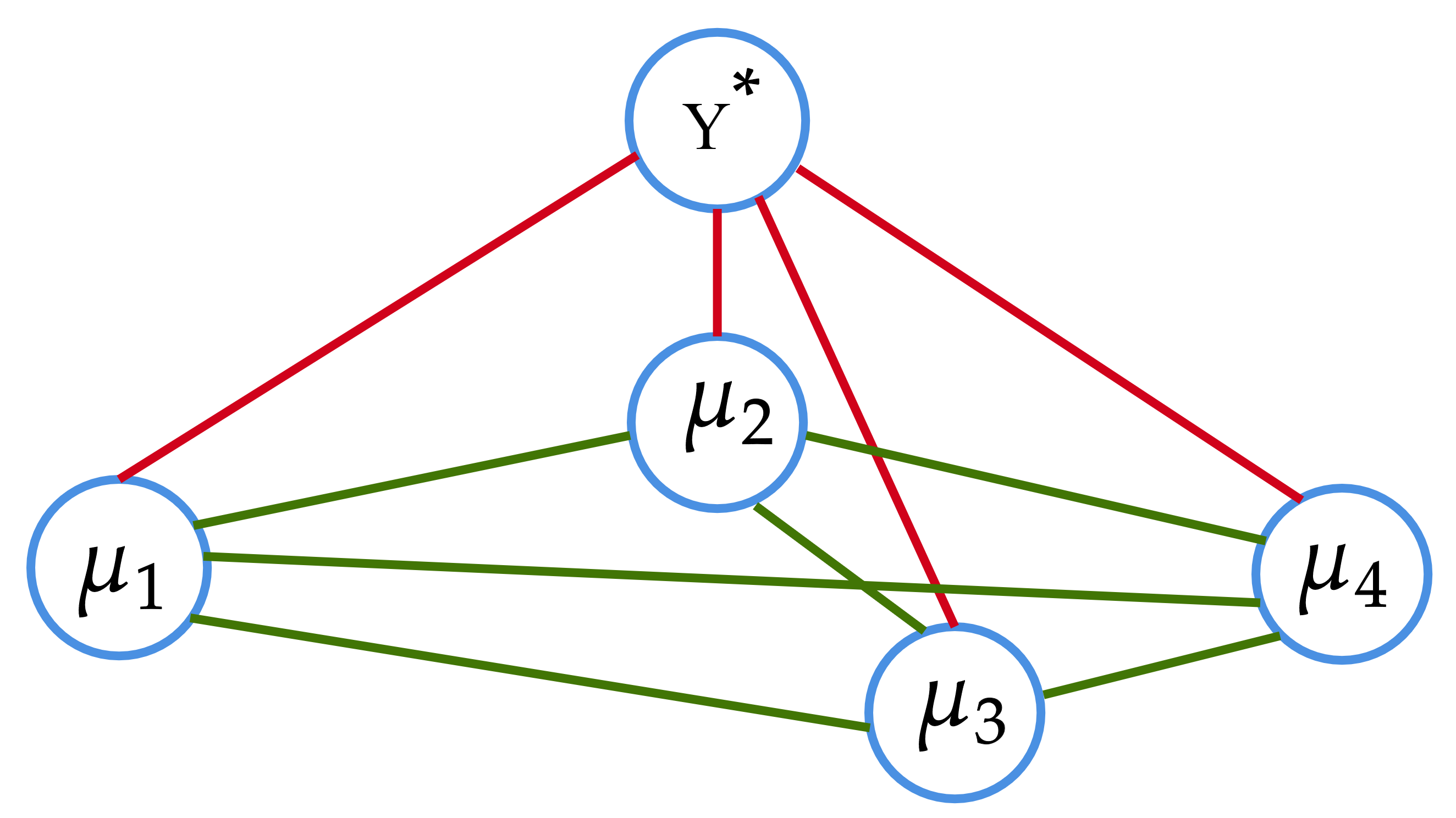}} 
    \caption{\textbf{Computational graphs} of different aggregation strategies. (a) Conditional-Independent based aggregation where there is no interaction between the local experts. (b) An aggregation based on the conditional dependency between local experts where their interactions have not been excluded.}
    \label{aggregation_graphs}
    %\vskip -0.2in
\end{figure}

Figure \ref{aggregation_graphs} depicts the computational graphs of both strategies. Figure \ref{aggregation_graphs}(a) reveals the aggregation based on conditional independence assumption between experts $\{\mu_1,\mu_2,\mu_3,\mu_4 \}$. It means two local experts $mu_i$ and $mu_j$ are connected only via the target variable $y^*$, i.e. $mu_i \indep mu_j \mid y^*$. However, this assumption is often violated in realistic conditions and the aggregation can lead to a sub-optimal solution. On the other hand, Figure \ref{aggregation_graphs}(b) represents an aggregation with dependent experts where the interactions between experts show the dependencies. 

\section{Proof of Proposition \ref{prop_blup}} \label{App.proof}
\begin{proof}
The proof is straightforward. We need to show that 
$Var(y-\beta \mu^*) -Var(y-y_A^*)$ is positive semi-definite for all linear unbiased predictors $\beta \mu^*$. To do that, we extend the $Var(y-\beta \mu^*)$:

\begin{align*}
&{}Var(y-\beta \mu^*)= Var(y-y^*_A + y^*_A -\beta \mu^*) = Var(y-y^*_A) + var(y^*_A-\beta \mu^*) + 2cov(y-y^*_A, y^*_A- \beta \mu^*).
\end{align*}

Now, we show $cov(y-y^*_A, C\mu^*)=0, \forall{C}$.  
\begin{align*}
&{}cov(y-y^*_A, C\mu^*) = cov(y,C\mu^*) - cov(y^*_A, C \mu^*) = \Sigma_{y^* \mu^*} C^T - \Sigma_{y^* \mu^*} \Sigma_{\mu^*}^{-1} \Sigma_{\mu^*} C^T  = 0.
\end{align*}
Therefore, $cov(y-y^*_A,y^*_A- \beta\mu^*)=0$ where in this case, $C=\Sigma_{y^*\mu^*}^T \Sigma_{\mu^*\mu^*}^{-1} - \beta$. It means 
\begin{align*}
&{}Var(y-\beta \mu^*)-Var(y-y^*_A) = var(y^*_A-\beta \mu^*)  \geq 0
\end{align*}
because $var(y^*_A-\beta \mu^*)$ is positive semi-definite variance matrix. 
\end{proof}

\section{Computational cost of EMGGM and NPAE} \label{App.complexity}
Both EMGGM and NPAE use  dependent experts. However, there are two major differences between them. First, NPAE needs all training and test data points during aggregation. Let $\Gamma_i =k^T_{i*}(K_i + \sigma^2 I)^{-1}$. For a test point $x^*$, the pointwise covariance between experts i and j in NPAE, $K(x^*)_{ij}$, can be extended using \eqref{eq:2} as     
\begin{align*}
K(x^*)_{ij}=cov(\mu^*_i(x^*), \mu^*(x^*)_j)=Cov(\Gamma_i y_i,\Gamma_j y_j) \notag  
   =\Gamma_i Cov(y_i,y_j) \Gamma_j^T = \Gamma_i k(x_i,x_j) \Gamma_j^T.
\end{align*}

Therefore, all auto-covariance $k(x_i,x_i)$ and cross-covariance $k(x_i,x_j)$ matrices are required for NPAE aggregation which raises the storage costs. 

Second, both aggregation methods have a $\mathcal{O}(M^3)$ calculation in each iteration, the inverse of $M \times M$ matrix in NPAE and GLasso in the proposed method. NPAE should do this costly calculation at each test point and therefore it is not efficient for large data sets. However, the proposed model can converge after a small number of iterations. When $R \ll n_t$, the proposed method is much faster than NPAE. Although the conventional GLasso for network learning is a costly method $\mathcal{O}(M^3)$, there are newer faster methods to learn a GGM that can be used instead of the GLasso, see \cite{wang_2016, Xu2017,zhang_2020}. For instance, the FST model \cite{zhang_2020} reduces the computational complexity of sparse Gaussian Graphical Model to a much lower order of magnitude $\left(\mathcal{O}(M^2)\right)$. 

\section{Gaussian Assumption} \label{App.Gaussian}
The normality assumption for joint distribution is not a restrictive assumption. In practice, we can relax this assumption and consider random variables without resorting to multi-dimensional Gaussian distribution. As a semiparametric generalization for continuous variables, authors in \cite{Liu2009, Lafferty} introduced the nonparanormal graphical model where it is assumed that the variables follow a Gaussian graphical model only after some unknown smooth monotone transformations on each of them. \cite{Mulgrave_2020} considered Bayesian inference in nonparanormal graphical models by putting priors on the unknown transformations through a random series based on B-splines. 

On the other hand, nonparametric methods can be used for functional graphical models. Authors in \cite{li} and \cite{Solea_2021} exerted additive conditional independence and functional principal components to learn a graphical model when observations on vertices are functions. This gives the result that the proposed strategy can be considered as a general ensemble model, and not only for the local approximation GPs.

\section{Latent Variable GGMs} \label{App.LVGGM}
Latent variable GGMs (LVGGMs) are used to estimate the distribution of the observed variables with respect to some latent variables. GGMs with latent variables have been widely considered over the past decade. Authors in \cite{Chandrasekaran} proposed \textit{Low-Rank Plus Sparse Decomposition} (LR+SD), a regularized maximum likelihood approach to estimate $\Omega$ via convex optimization. The precision matrix in LR+SD contains two terms: sparse structure $\Omega_{\mu^*}$ and the low-rank terms $L^*=\Omega_{\mu^*y^*} \Omega_{y^* y^*}^{-1} \Omega_{y^*\mu^*}$. The precision matrix in this form is $\Omega=\Omega_{\mu^* \mu^*}-  L^*$. The log-likelihood can be expressed in terms of the $S_{\mu^* \mu^*}$, $\Omega_{\mu^* \mu^*}$, and $L^*$: 
\begin{equation} \label{log_lik_lvggm}
\mathcal{L}(\Omega_{\mu^* \mu^*}, L^*; S_{\mu^* \mu^*}) 
=  log |\left(\Omega_{\mu^* \mu^*}-  L^* \right)| - trace\left(S_{\mu^* \mu^*} \left(\Omega_{\mu^* \mu^*}-  L^* \right)\right).   
\end{equation}

Essentially, it is a misspecified optimization problem because the precision matrix is the sum of two matrices. However, if $\Omega_{\mu^* \mu^*}$ is sparse and there are few latent variables, it is possible to decompose the precision matrix into its summands \cite{Candes, Chandrasekaran}. 
\begin{align}
\left(\hat{\Omega}_{\mu^* \mu^*}, \hat{L}^* \right)  &=  \underset{\Omega_{\mu^* \mu^*}, L^*  \in \mathcal{R}^{M\times M}}{\arg\min}  -\mathcal{L}(\Omega_{\mu^* \mu^*}, L^*; S_{\mu^* \mu^*}) + \lambda \left(\gamma \left\Vert \Omega_{\mu^* \mu^*} \right\Vert_1 + \left\Vert L^* \right\Vert_* \right) \label{eq:lvggm}   \\
   &{} \text{such that \;\;  }  \Omega_{\mu^* \mu^*}-  L^* \succ 0 ,\; L^* \succeq 0  
\end{align}

such that $\Omega_{\mu^*}-  L^* \succ 0 ,\; L^* \succeq 0$. Here, $\lambda>0$ and $\gamma > 0$ are tuning parameters for sparsity and low rankness, and $\left\Vert L^* \right\Vert_*$ denotes the nuclear norm of $ L^*$ (i.e. the sum of its singular values).

To speeding up the LR+SD model, \cite{Xu2017} proposed a non-convex optimization model and showed that it is orders of magnitude faster than the convex relaxation-based methods. Author in \cite{yuan} proposed a direct approach via Expectation-Maximization algorithm which converts LR+SD model to a conventional GGM. Here, we modified this approach and proposed EMGGM. 

However, LR+SD model has been developed to estimate the marginal distribution of observed variables, i.e. $p(\mu^*)= \int p(\mu^*,y^*)dy^*$, while the desired predictive distribution is the conditional distribution of $y^*$ given local experts' predictions $p(y^*|\mu^*)$. Hence, further work could consider the modified form of the log-likelihood in \eqref{log_lik_lvggm} and the convex optimization in \eqref{eq:lvggm} to estimate the aggregate estimator $y^*_A$ in \eqref{aggregated_estimator} via a convex or non-convex optimization problem.

\end{appendices}

\end{document}